\documentclass[sigconf]{acmart}

\setcopyright{none}
\renewcommand\footnotetextcopyrightpermission[1]{}

\usepackage{amsmath}
\usepackage{amsthm}

\usepackage{tikz}
\usetikzlibrary{positioning,shadows,shapes,arrows.meta}

\newtheorem{theorem}{Theorem}

\newtheorem{proposition}[theorem]{Proposition}

\newtheorem{definition}{Definition}

\newcommand{\BibTeX}{B\kern-.05em{\sc i\kern-.025em b}\kern-.08em\TeX}

\newcommand{\defin}{=_{\textit{def}} }

\newcommand{\atm}{\mathit{Atm}}

\newcommand{\val}{\mathit{Val}}

\renewcommand{\phi}{\varphi}

\newcommand{\putaway}[1]{}

\newcommand{\cb}{\mathit{CB}}

\newcommand{\outcome}{\mathit{o}}

\newtheorem{remark}{Remark}%

\begin{document}

\title{Rule-based Classifier Models }

\author{Cecilia Di Florio}
\email{cecilia.diflorio2@unibo.it}
\orcid{0000-0002-8927-7414}
\affiliation{
  \institution{University of Bologna, University of Luxembourg}
  \city{}
  \country{}}

\author{Huimin Dong}
\email{Huimin.dong@tuwien.ac.at}
\orcid{0000-0002-4951-0111}
\affiliation{
  \institution{TU WIEN}
\city{}
  \country{}}

  \author{Antonino Rotolo}
\email{antonino.rotolo@unibo.it}
\orcid{0000-0001-5265-0660}
\affiliation{
  \institution{University of Bologna}
  \city{}
  \country{}}

\begin{abstract}
We extend the formal framework of classifier models used in the legal domain. While the existing classifier framework characterises cases solely through the facts involved, legal reasoning fundamentally relies on both facts and rules, particularly the \emph{ratio decidendi}. This paper presents an initial approach to incorporating sets of rules within a classifier. Our work is built on the work of Canavotto \emph{et al.} (2023), which has developed the rule-based reason model of precedential constraint within a hierarchy of factors. We demonstrate how decisions for new cases can be inferred using this enriched rule-based classifier framework. Additionally, we provide an example of how the time element and the hierarchy of courts can be used in the new classifier framework
\end{abstract}

\maketitle
\pagestyle{plain}

\section{Introduction}
There are numerous concerns about using machine learning classifiers to predict judicial outcomes. These concerns include the uncertainty regarding the accuracy and normative validity of the classifiers' predictions, as well as a lack of transparency and explainability in their decision-making processes. To address these issues, symbolic methods are required to formally verify the robustness of machine learning algorithms used in predictive justice. 

Machine learning algorithms in predictive justice perform \emph{case-based reasoning} (CBR), in the sense that they provide the outcomes of new cases on the basis of previous cases. Based on this intuition, \cite{liu2022modelling} investigated a correspondence between existing formal models of legal CBR in  \cite{Horty2011RR} and binary input classifier models proposed in \cite{LiuLoriniJLC}.  The classifiers framework for legal reasoning was then expanded in \cite{lail, FlorioLLRS23, PrecedentsClash}.  In particular in \cite{PrecedentsClash}, classifier models were enriched to account for two fundamental elements  in legal practice:  the temporal element and the hierarchical structure of courts within the  legal system under consideration. These two elements played a crucial   role  for refining the notion of constraining/binding precedent and for solving conflicts among binding precedents.
We believe the classifier models proposed in \cite{PrecedentsClash} can be further refined. In these framework, every  case  is essentially described  in terms of the facts that appear in the case and of the court that assessed the case. Additionally in \cite{PrecedentsClash}, it is assumed that  the decision making process for each new case is a single-step process that, on the basis of the binding precedents, leads directly from the facts of case  to its outcome. In legal practice, however, cases are not defined solely in terms of facts, but also by the \emph{ratio decidendi} of the case, which is the reason for the courts decision. The \emph{ratio decidendi} is generally binding on lower courts and later judgments\cite{LexisNexis2}. In some formal models of legal CBR \cite{Horty2011RR}, the \emph{ratio decidendi} is represented as a rule linking directly  specific facts of the case to the conclusion of the case. 
However, as suggested in \cite{branting, Canavotto1}, and recalled in \cite{Canavotto1},  the reasoning pattern in  judicial opinions  is often based on inferences linking portions of the facts of the case to intermediate conclusions supported by those facts. In \cite{branting} it is argued that also these intemediate rules, referred to as ``precedents constituents,'' should constrain in later cases. 
Different approaches have been proposed in legal CBR about how portions of precedents may condition future decisions.  For instance, in both the CATO \cite{CATO} and IBP\cite{IBP} systems, the basic factors of a case are organized into issues. In CATO  however, an extra layer of intermediate/abstract factors exists, positioned between the issues and the base-level factors. In \cite{BenchCapon1, BenchCapon2}, it is argued that only decisions regarding the issues (and not the intermediate factors) can be used to constrain future cases. In \cite{Canavotto1,Canavotto2}, instead, it is argued that decisions concerning intermediate factors can also impose constraints on subsequent cases.  

 The aim of this article is not to argue for or against different approaches to precedential constraint, but to develop a rule-based classifier model that integrates rules into the existing classifier framework.  We want to enrich the classifier models so that the states of the classifiers are defined not only in terms of facts, but also in terms of \emph{sets} of rules. Our long-term goal is to show how the framework in \cite{PrecedentsClash} can be combined with classifiers that take different rule systems into account. In this work, as a starting point, we build classifier modeled on  the CBR framework provided in \cite{Canavotto1}. The work in~\cite{Canavotto1} develops the reason model of  precedential constraint \cite{Horty2011RR}  in the context of a factor hierarchy, so that the final decision of a case depends on the decisions of intermediate concerns and on the rules used in those decisions.

The paper is structured as follows. Section \ref{sect1:canavotto} reviews the framework in \cite{Canavotto1}. Section \ref{section:classifiers} enriches the framework of classifiers model with sets of rules, starting from the framework in \cite{Canavotto1}, and demonstrates how a decision making process can be defined within this new framework. Section~\ref{section:binding} illustrates the benefits of integrating the  framework from \cite{PrecedentsClash} with the new classifier models. Section~\ref{sec:conclusion} concludes.~\footnote{\;This paper is an version of a short paper accepted to ICAIL 2025}

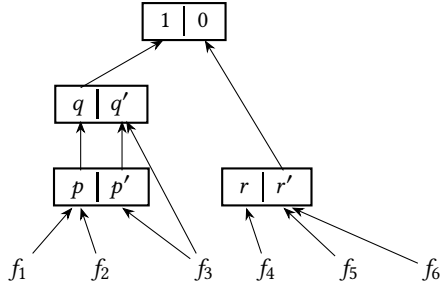
\begin{figure}
    \centering
    \begin{tikzpicture}[scale=0.55,
    every node/.append style={thick, rectangle,minimum size=1.5pt, inner sep=1.5pt},>={Stealth[scale=1]} ]
    
    \node[draw] (4) at (2,3) {\begin{tabular}{c|c}
        1 & 0 
    \end{tabular}};
    
    \node[draw] (3) at (0,1) {\begin{tabular}{c|c}
        $q$ & $q'$ 
    \end{tabular}};
    
    \node[draw] (21) at (0,-1) {\begin{tabular}{c|c}
        $p$ & $p'$ 
    \end{tabular}};

    \node[draw] (22) at (4,-1) {\begin{tabular}{c|c}
        $r$ & $r'$ 
    \end{tabular}};

    \node (f1) at (-2,-3) {$f_1$};
    \node (f2) at (0,-3) {$f_2$};
    \node (f3) at (2.5,-3) {$f_3$};
    \node (f4) at (4,-3) {$f_4$};
    \node (f5) at (6,-3) {$f_5$};
    \node (f6) at (8,-3) {$f_6$};

\path[->] (f1) edge (-0.7,-1.5);
\path[->] (f2) edge (-0.5,-1.5);
\path[->] (f3) edge (0.5,-1.5);
\path[->] (f3) edge (0.6,0.6);
\path[->] (0.5,-0.6) edge (0.5,0.6);
\path[->] (-0.5,-0.6) edge (-0.5,0.6);
\path[->] (-0.5,1.4) edge (1.5,2.55);

\path[->] (f4) edge (3.5,-1.5);
\path[->] (f5) edge (4.4,-1.5);
\path[->] (f6) edge (4.6,-1.5);
\path[->] (4.4,-0.6) edge (2.5,2.55);
\end{tikzpicture}
    \caption{An abstract example from~\cite{Canavotto1}}
    \label{fig:exampleH}
\end{figure}

\section{Hierarchical Reason Model}\label{sect1:canavotto}
In this section we will recall some notions introduced in \cite{Canavotto1}, adapting the notation and terminology to our purposes.

The work in \cite{Canavotto1} addresses the problem of precedential constraint with respect to a single top-level issue. With respect to the issue the judge can decide in favor of the plaintiff or in favor of the defendant. From now on, we will write $0$ for a decision in favor of the defendant and $1$ for a decision in favor of the plaintiff; the set of these two decisions will be denoted as $Dec=\{0,1\}$.
In addressing the case, the judge starts by considering a set of base factors (factual facts), denoted as $Atm_0$. Given these base factors, the judge may decide whether some intermediate factors apply in the case within a multi-step procedure~\cite{Canavotto1}. The set of intermediate factors is denoted as $Int$. Following \cite{Canavotto1}, we will assume that if $p$ is an intermediate factor in $Int$, then its contrary, noted $p'$, is also in $Int$. On the basis of the intermediate factors, some further intermediate factors could be inferred.  This step-by-step procedure is performed until a decision for the top issue ($0$ or $1$) is reached.  

Recall that in the single-step reason and the result models by Horty~\cite{Horty2011RR}, the factors that may appear in a case are partitioned into factors favoring a decision for the plaintiff ($1$) and factors favoring a decision for the defendant ($0$). Similarly, in the multi-step model~\cite{Canavotto1}, base factors and intermediate factors may favor the decision for an intermediate factor $p$ or its contrary; or they may favor the decision for $0$ or $1$. We define $X\subseteq \atm_{0}\cup Int$ a factual situation. 

As in \cite{Canavotto1}, we use $t,u, v$ for variables ranging over elements in $\atm_{0}\cup Int\cup Dec$. Given $t\in Int\cup Dec$, we denote $\overline{t}$ its opposite. More precisely,  if $t=p\in Int$, $\overline{t}= p'$, if $t=p'$ then $\overline{t}=p$. Also, if $t=0\in Dec$, $\overline{t}= 1$, if $t=1 \in Dec$ then $\overline{t}=0$. Finally, $\overline{\overline{t}}=t$.

A hierarchy of factors is used to specify which intermediate factor/final decision is favored by each base factor/intermediate factors~\cite{Canavotto1}, related to a given problem-domain. In the following we will mark by $*$ the definitions we took from \cite{Canavotto1}, sometimes adapting the notation to our purposes. 

\begin{definition}[* Factor link and Factors Hierarchy]
Let $t\in \atm_{0}\cup Int$ and $u\in Int\cup Dec$. A factor link is a statement of the form $t\to u$
indicating that the presence of the factor $t$ in some situation directly favors a decision that $t$ holds as well, or simply that $t$ directly favors $u$. A factor hierarchy is a set H of factor links.
\end{definition}

\begin{remark}\label{remark}
 We assume that, given a hierarchy $H$, it is not possible to have factors $t,u$ such that both $t\to u$ and $\overline{t}\to u$ are in $H$ (a factor and its opposite cannot favor the same factor); and, further, it is not possible to have $t,u$ such that both $t\to u$ and $t\to \overline{u}$ are in $H$ (a factor cannot favor a factor and its opposite). Finally we assume $H$ to be acyclic.
\end{remark}

\begin{example}\label{ex:Hierarchy2}
From now on, we refer to a  hierarchy of factors defined in \cite{Canavotto1} and represented in Figure~\ref{fig:exampleH}. We call the hierarchy $H_{ex}$. Here,  $\atm_{0}= \{f_1, f_2,f_3, f_4,f_5, f_6\}$, $Int= \{p,p',r,r',q,q'\}$, and $Dec=\{0,1\}$.
\end{example}

Given $t\in Int$ an intermediate factor, we use notation $t/\overline{t}$ to indicate an intermediate concern. Intuitively, the  concern represented by $t/\overline{t}$ is whether the intermediate factor $t$ or its opposite  $\overline{t}$ should hold in the  situation under consideration. Similarly, given $t\in Dec$  top-level decision, we define $t/\overline{t}$ as a top-level concern.  
A top-level concern is also called top-level issue.
The set of all the concerns (intermediate and top-level) is $Concern$. 

\begin{definition}[*Favoring factors]
Let $t\in \atm_{0}\cup Int$ and $u\in Int\cup Dec$. We say that $t$ favors $u$, if in $H$ there is a direct edge $t\to u $ or  there is a direct edge $\overline{t}\to \overline{u} $ . 
 We define $Facts^{u}$ the set of factors favoring $u$.
 \end{definition}

\begin{example}\label{ex:Canavotto1}
Consider $H_{ex}$ in Figure \ref{fig:exampleH}. $f_{1}$ is a factor in favor of $p$, since there is an edge $f_{1}\to p$; namely $f_{1}\in Facts^{p}$. While, $r$ is a factor in favor of $1$, since there is an edge $r' \to  0$;  so, $r\in Facts^{1}$. 
\end{example}

The idea in \cite{Canavotto1}, is that the concerns  should be solved sequentially, following their position in the hierarchy:  concern with lower position  in the hierarchy should be solved before and, on the basis of the solutions of the lower concerns, concerns with higher position in the hierarchy can be solved. Namely, concerns should be solved following the order provided  by their \emph{degree}.

\begin{definition}[*Degree of factors and concerns]\label{def:degree}
Let $t\in  \atm_{0}\cup Int \cup Dec$. We define the degree of $t$, denoted $degree(t)$, as follows:
\begin{itemize}
\item $degree(t)= 0$ iff $t\in  \atm_{0}$ 
\item $degree(t)= 1+max\{degree(u)\mid u \in Facts^{t}\}$.
\end{itemize}
Let $t/\overline{t} \in Concern$. We define the  degree of $t/\overline{t}$ as $degree(t/\overline{t})= degree(t)= degree(\overline{t})$.
\end{definition}

\begin{remark} \label{prop:dg>1}
Notice that if $t/\overline{t}$ is a concern, then  $degree(t/\overline{t})\geq 1$: indeed, in this case $t\in Int\cup Dec$ and so from Definition~\ref{def:degree}, we know that $degree(t)=0$ iff $t\in \atm_{0}$. 
\end{remark}

But, as pointed out in \cite{Canavotto1}, not all concerns should be addressed in every situation: when dealing with a fact situation $X$ , only those concerns that are actually \emph{raised} by $X$, should be addressed. 

\begin{definition}[*Concerns raised by a fact situation]\label{def:concernFactsituation}
Let $H$ be a hierarchy and $X\subseteq \atm_{0}\cup Int$. The set of concerns that are raised by $X$ is $Concern(X)=\{t/\overline{t} \in Concern: X\cap (Facts^{t}\cup Facts^{\overline{t}})\neq \emptyset \}$. 
\end{definition}

\begin{definition}[*Concerns of degree n raised by a fact situation]
 Let $H$ be a hierarchy with $degree(0/1)=m$. Given $1\leq n\leq m$, the set of concerns of degree $n$ that are raised by $X$ is $Concern^{n}(X)=Concern(X) \cap \{t/\overline{t} : degree(t/\overline{t} )=n \}$.
\end{definition}

In assessing each concern raised by $X$ the court will determine not only the outcome of the decision for the concern but also the \emph{reason} for that decision. In a certain sense, we may say that in each decision on a concern, a rule is established: the antecedent of the rule is the \emph{reason} for the decision, and the conclusion of the rule is the outcome related to the decision.  In the context of a hierarchy of factors, the rules used in a decision must be admissible wrt the hierarchy: a rule cannot introduce  links among factors that cannot be extracted from the hierarchy. 

\begin{definition}[Rule admissible in $H$]\label{def:admissibleRule}
Let $H$ be a hierarchy. A rule  admissible in $H$ is a statement of the form $r: U\to t$, where $t\subseteq Int\cup Dec$ and $U\subseteq Facts^{t}$. In addition, $Antecedent(r) = U$ is the antecedent of the rule and $Conclusion(r ) = t$ its conclusion.
\end{definition}

 We denote $Rules$ the set of all admissible rules in $H$. 
From the previous definition it follows that the antecedent of an admissible rule $r$ can be a set of base factors or intermediate factors; while the conclusion of an admissible rule $r$ is a single element that can be either an intermediate factor or an element in $Dec$.

\begin{definition}[*Rule applicable in a fact situation]
A rule is applicable to a base fact situation $X\subseteq \atm_{0}$ iff $Antecedent(r)\subseteq X$. 
\end{definition}

\begin{definition}[*Decision]\label{def:decision}
A decision is a triple $d = (X,r, t)$ where $X\subseteq \atm_{0}\cup Int$ is some fact situation, $r$ is applicable to $X$, and $Conclusion(r) = t$. We define $Outcome(d)= Conclusion(r)$.
\end{definition}

\begin{definition}[* (Complete) resolution]\label{def:completeresolution}
Given $t/\overline{t}\in Concern$, a resolution related to $t/\overline{t}$ is a decision
$d = (X,r, Conclusion(r))$, where $Conclusion(r)\in\{t, \overline{t}\}$. Given $Cn\subseteq Concern$, a complete resolution of $Cn$ based on a fact situation  $X$ is a set containing, for each concern from $Cn$, exactly one resolution of that concern based on that fact situation $X$. 
\end{definition}

Finally, \cite{Canavotto1} defines a procedure to collect all resolutions of concerns in an opinion by starting from a factual situation and moving upward through the hierarchy of factors. 

\begin{definition}[*Opinion based on a fact situation]\label{def:OpinionCase}
Given a single-issue factor hierarchy H and a base-level fact situation $X \subseteq \atm_{0}$, an opinion based on $X$ in the context of $H$ is defined as an output of the following procedure:
\begin{itemize}
\item Input the fact situation $X$; 
\item Set (a) $Y_0 =X$, (b) $m = degree(0/1)$;
\begin{itemize}
\item  For $n=0$ to $m-1$,
\begin{itemize}
\item [(a)] Let $Res_{n+1}$ be some complete resolution of $Concern^{n+1}(Y_{n})$ based on the fact situation $Y_{n}$;
\item [(b)] Set $Y_{n+1} = Y_{n} \cup \{Outcome(d) : d \in Res_{n+1}\}$;
\end{itemize}
\item Output $op = (Res_{1},  ..., Res_{m} )$ as an opinion based on $X$.
\end{itemize}
\end{itemize}
\end{definition}

We can now introduce the notion of case and case bases. 

\begin{definition}[*Case and Case Base] \label{def:cb}
A case, in the context of a hierarchy $H$, is a triple  $c = (X,op,o)$, where $X\subseteq \atm_{0}$ is a base-level fact situation,  $op$ is an opinion based on $X$ in that hierarchy, and where $o\in \{0,1\}$ is the outcome supported by $op$. A case base is a set of cases. 
\end{definition}

\begin{remark}
Given a case $c = (X,op,o)$, by Definition~\ref{def:completeresolution}, we know that for each concern $t/\overline{t}$ there is atmost one decision related to $t/\overline{t}$ in the opinion of the case. 
\end{remark}

\begin{example} \label{ex:CasebaseCanavotto}
Consider the factual situations $X_{1}= \{f_2, f_3, f_4, f_5\}$ and $X_{2}= \{f_1, f_3,  f_4, f_5, f_6\}$. We provide two examples of cases
\begin{itemize}
\item $\tilde{c}_{1}= (X_{1},op_{1}, 1)$, where , $op_{1}=(Res_1, Res_2, Res_3)$
\begin{itemize}
\item $Res_{1}= \{(X_1, \{f_{2}\}\to p, p), (X_{1}, \{f_{4}\}\to r, r)\} $; 
\item $Res_{2}=\{(Y_{1}, \{p\}\to q, q)\} $,  $Y_{1}=X_{1}\cup\{p,r\}$;
\item $Res_{3}= \{(Y_{2}, \{q\}\to 1, 1)\}$ , with $Y_{2}=X_{1}\cup\{p,q,r\}$.
\end{itemize}
\item $\tilde{c}_{2}= (X_{2},op_{2}, 1)$, where ,  $op_{2}=(Res_1, Res_2, Res_3)$.
\begin{itemize}
\item $Res_{1}= \{(X_2, \{f_{1}\}\to p, p),  (X_{2}, \{f_{5},  f_{6}\}\to r', r')\} $; 
\item $Res_{2}=\{(Z_{1}, \{p\}\to q, q)\} $, with $Z_{1}=X_{2}\cup\{p,r'\}$;
\item $Res_{3}= \{(Z_{2}, \{r'\}\to 0, 0)\}$ , with $Z_{2}=X_{2}\cup\{p,r',q\}$.
\end{itemize}
\end{itemize}
$\tilde{CB}=\{ \tilde{c}_1, \tilde{c}_2\}$ is a case base.
\end{example}

\cite{Canavotto1} imports the notion of precedential constraint described in the reason model \cite{Horty2011RR} into the hierarchical model described so far. In reason model, which focus directly on the $0/1$ concern (i.e. on whether a case should be decided in favor of the plaintiff or the defendant in a single-step procedure), the precedential constraint is based on the idea that a court's decision establishes a priority ordering among the reasons for the plaintiff and the reasons for the defendant. Subsequent courts are then constrained to issue decisions that are consistent with the priority orderings established by previous cases. The same idea can be applied to the hierarchical models. The main difference is that now each case in the case base establishes several priority orderings, one  for every concern that was addressed in the case. Because of this, \cite{Canavotto1}  firstly introduces the notion of priority ordering among reasons pertaining to a concern and derived by a decision

\begin{definition}
In the context of a hierarchy $H$, let $t/\overline{t}$ be a concern, $U\subseteq Facts^{\overline{t}}$,  $V\subseteq Facts^{t} $, and $d = (X,r, t)$ be a decision. Then the relation $ <^{t/\overline{t}}_{d}$ representing the priority ordering on reasons relative to $t/\overline{t}$ and derived from $d$
 is defined by stipulating that $U <^{t/\overline{t}}_{d} V$ if and only if $U \subseteq X$ and $Antecedent(r)\subseteq V$.
\end{definition}

Then, all the decisions that appear in the opinion of a case are  merged into a single set.

\begin{definition}[*Merge of an opinion]
Let  $m = degree(0/1)$. Given an opinion $op = (Res_{1}, Res_{2}, ..., Res_{m})$, the merging of $op$ is  $Merge(op) = \{Res_i : 1 \leq i \leq  m\}$.
\end{definition}

A priority ordering among the reasons relating to a concern derived from a case is then defined. This is the priority ordering among the reasons induced by the decision related to the concern that appears in the opinion of the case.  

 \begin{definition}[*Priority ordering relative to concern derived from case] \label{def:priorityConcern}
 In the context of a hierarchy, let $t/\overline{t}$ be a concern, $U\subseteq Facts^{\overline{t}}$, $V\subseteq Facts^{t} $, and
$c = (X,op,o)$ be a case. Then, the relation $ <^{t/\overline{t}}_{c}$ representing the priority 
ordering on reasons relative to $t/\overline{t}$ and derived from $c$ is defined by stipulating that $U <^{t/\overline{t}}_{c} V$  if and only if there is some decision $d\in Merge(op)$ such that $U <^{t/\overline{t}}_{d} V$ 
\end{definition}

The priority ordering among the reasons in a  case base is then defined from the preference ordering among the reasons induced by the cases that are in the  case base.  

\begin{definition}[*Priority ordering relative to concern derived from case base]\label{def:pref case base}
 In the context of a hierarchy, let $t/\overline{t}$ be a concern, $U\subseteq Facts^{\overline{t}}$, $V\subseteq Facts^{t} $, and $\cb$ be a case base. Then the relation $ <^{t/\overline{t}}_{\cb}$ representing the priority ordering on reasons relative to $t/\overline{t}$ and derived from $\cb$ is defined by stipulating that $U <^{t/\overline{t}}_{CB} V$  if and only if there is some case $c\in \cb$ such that $U <^{t/\overline{t}}_{c} V$ 
\end{definition}

We can finally say that a case base is inconsistent wrt a concern when the case base provides conflicting information on the priority of reasons wrt the concern. 

\begin{definition}[(In)cosistency wrt a concern]\label{def:concerninconsistency}
Given a concern $t/\overline{t}$, a case base $\cb$ is inconsistent wrt $t/\overline{t}$, if there are $U\subseteq Facts^{\overline{t}}$ and $V\subseteq Facts^{i} $ such that both  $U <^{t/\overline{t}}_{CB} V$ and $V<^{t/\overline{t}}_{CB} U$. We say $\cb$ is consistent wrt $t/\overline{t}$ if it is not inconsistent wrt $t/\overline{t}$.
\end{definition}

\begin{definition}[(In)cosistency]\label{def:concerninconsistency2}
We say that a case base $\cb$ is inconsistent if there is a concern $t/\overline{t}$ s.t. $\cb$ is inconsistent wrt $t/\overline{t}$. A case base is consistent if it is not inconsistent.
\end{definition}

\begin{example}\label{ex:CasebaseCanavotto2}
It can be verified that $\tilde{\cb}$ in Example \ref{ex:CasebaseCanavotto} is consistent --- no conflicting priority orderings can be  derived by $\tilde{c}_{1}$ and $\tilde{c}_{2}$. 
\end{example}

\section{Generalizing Classifier Models: From Single to Many Decisions}\label{section:classifiers}

In this section, we want to show how the hierarchical models presented in the previous section can be represented in the framework of classifier models introduced in~\cite{LiuLoriniJLC} and adapted to legal case-based reasoning in~\cite{liu2022modelling, lail,FlorioLLRS23,PrecedentsClash}. In these classifier models, there are essentially two components: a set of states and a classification function that associates each state with a possible outcome --- where the outcome depends on the classification task. The states are described in terms of input features/atoms. In the classifier models for the legal domain, the input features can be the facts/factors that may appear in a case, while the outcome can be $0$ (decision in favor of the defendant), $1$ (decision in favor of the plaintiff) or $\sf{?}$ (absence of decision).  We want to modify the classifier framework so that not only factors but also rules can be considered when describing cases. More precisely, a state of the classification model will be a set of facts together with a set of rules, used to describe the case.  In this paper we will focus on the specific sets of rules; we call these sets solutions.\footnote{It would be more appropriate to call them opinion; we use the term solution to avoid confusion with opinion in \cite{Canavotto1}}.

\begin{definition}[Rule related to a concern]\label{def:ruleConcern}
 Given $t/\overline{t}\in Concern$, we say that a rule $r\in Rules$ is related to concern $t/\overline{t}$ iff $Conclusion(r) \in \{t,\overline{t}\}$.
\end{definition}

\begin{definition}[Applicable solution]\label{def:sol}
Given a hierarchy $H$ and $X\subseteq \atm_{0}$, a solution in $H$ applicable to $X$ is any subset $sol \subseteq Rules $, when: 
\begin{enumerate}
\item  $\forall r\in sol$: if $p=Antecedent(r)\in \atm_{0}$ then $p\in X$;
\item for all rules $r\in sol$, it is the case that $Antecedent(r)\subseteq X \cup \bigcup_{r'\in sol}  Conclusion(r')$; and 
\item[(3)] For all $t/\overline{t}\in Concern_{H}(Conclusion(r'))\cup  Concern_{H}(X)$ with $r'\in sol$, there is a unique $r\in sol$ related to this concern. 
\end{enumerate}
\end{definition}

Condition 2) states that in a solution there are not rules whose antecedent is not in $X$ or  that cannot be derived from other rules in  the solution. Condition 3) states that for each concern raised by the conclusion of the rules in the solution or by the factual situaion in $X$, there is a rule related to the concern.

Given $X\subseteq \atm_{0}$, we let $Sol_{X}$ be the set of all applicable solutions to $X$ and $Sol\defin \{Sol_{X}\mid X\subseteq \atm_{0}\}$ be the set  of all solutions in $H$. From the previous definition, it follows that, in a solution $sol$, there cannot be two conflicting rules: namely, there cannot be two rules $r,r'\in Sol$ such that $Conclusion(r)=t$ and $Conclusion(r')=\overline{t}$. Given $sol\in Sol$, for each concern $t/\overline{t}\in Concern$, we denote $rule_{sol}(t/\overline{t})$ as the unique  $r\in sol$ related to the concern $t/\overline{t}$, if it exists (we know it is unique because of third property in above definition).

We can now introduce the framework of rule-based classifer models.

\begin{definition}[Rule-based Classifier Model] \label{def:RCM}
    A rule-based classifier model (RCM) is a triple  $C = (S, H, f)$ where:
    \begin{itemize}
        \item $S=\{(X,sol)\mid X\subseteq \atm_{0}, sol\in Sol_{X}\} \subseteq 2^{Atm_0}\times Sol$;
        \item  H is a hierarchy;
        \item $f: S \to Val$ is a decision function, where $Val=\{0,1,?\}$. 
    \end{itemize}

\end{definition}
In a RCM, each state of the model $s=(X, sol)\in S$ contains a factual situation $X$ with an applicable solution $sol$. The classification function maps each state into a possible value, namely $\{ 0,1,?\}$. So, for each $s\in S$, we can have either that $f(s)=\outcome$ with $\outcome\in\{0,1\}$ and so $s$ represents a precedent case; or we can have that $f(s)=?$, and so $s$  is an unassessed/new case. 
Furthermore, we will impose the following constraint on the rule-based classifier models:
\begin{align}
&\forall s= (X, sol)\in S:  f(s)=? \Rightarrow sol = \emptyset  \tag{C1}\label{cond:1}\\
&\forall s= (X, sol)\in S: f(s)\neq ? \Rightarrow f(s)= Conclusion(rule_{sol}(0/1))\tag{C2}\label{cond:2}
\end{align}
With (\ref{cond:1}) we require that there is no solution for the undecided cases. With (\ref{cond:2}), we require that the classification of the state is consistent with a solution in the state. 

\begin{example}\label{ex:classifier}
We refer to the hierarchy in Figure \ref{fig:exampleH}.
 Let $S \subseteq 2^{Atm_0}\times Sol$. Let $f: S\to Val$ s.t:

\begin{itemize}
\item $f(s_{1})=1$, where  $s_{1}=(X_{1}, sol_{1})$,  with 
$X_{1}= \{f_2, f_3, f_4, f_5\}$,  $sol_{1}= \{\{f_2\}\to p, \{f_4\}\to  r  , \{p\}\to q, \{q\}\to 1\}$.
\item $f(s_{2})=0$, where $s_{2}=(X_{2}, sol_{2})$, with  $X_{2}= \{f_1, f_3, f_4, f_5, f_6\}$, $sol_{2}= \{\{f_1\}\to p, \{f_5,f_6\}\to  r'  , \{p\}\to q, \{r'\}\to 0\}$. 
\item $f(s)=?$, for all other $s\in S$.
\end{itemize}
We define the classifier model $C_{ex}= (S, f)$. In the following we will focus on the new state $s_{3}=(X_{3}, sol_{3})$ with $X_{3}= \{f_{1},f_2, f_3, f_4, f_5, f_6\}$, $sol_{3}=\emptyset$ and the classification $f(s_{3})=?$. 
\end{example}

We now underline the relation between the rule-based classifier models and case bases in Definition~\ref{def:cb}. Namely, we can define a way to translate a rule-based classifier model into a case base. Firstly, we show how a state can be translated to a case as defined in Definition \ref{def:cb}.

\begin{definition}[Case obtained from a state]\label{casefromstate}
Let $H$ be a hierarchy, $m=degree(0/1)$, and $s=(X, sol)\in S$ with $f(s)\neq ?$. The case obtained from $s$ is $c=(X, op(s), f(s))$ such that $op(s)=(A_1, A_2,..., A_m)$, where, for all $n\in \{1...,m\}$, $A_{i}$ is defined as follows. 
\begin{itemize}
\item Define $Y_0 =X$;
\item For $n=0$ to $m-1$,
\begin{itemize}
\item Set $A_
{n+1}=\bigcup_{ degree(t/\overline{t})= n+1}\{(Y_{n},r, Conclusion(r))\mid r=rule_{sol}(t/\overline{t})\} $
\item Set $Y_{n+1} = Y_{n} \cup \bigcup_{ degree(t/\overline{t})= n+1} \{Conclusion(r) \mid r=rule_{sol}(t/\overline{t})\}$
\end{itemize}
\end{itemize}
\end{definition}

\begin{example} \label{ex:opinionState}
Consider $s_{1}$ of example. We can define the case $c_{1}= (X_1, op(s_{1}), f(s_{1})=1)$, where $op(s_1)=(A_{1}, A_{2}, A_{3})$ is defined as follows:
\begin{itemize}
\item Set $Y_{0}=X_{1}$.
\item The concerns of degree 1 are $p/p'$ and $r/r'$. Hence, $A_{1}= \{(X_1, \{f_{2}\}\to p, p), (X_{1}, \{f_{4}\}\to r, r)\}$. 
\item Set $Y_{1}= Y_{0}\cup \{p, r\}$.
\item The concern of degree 2 is $q/q'$. Hence, $A_{2}= \{(Y_{1}, \{p\}\to q, q)\}$. 
\item Set $Y_{2}= Y_{1}\cup \{q\}$.
\item The concern of degree 3 is $0/1$ and so $A_{3}= \{(Y_{2}, \{q\}\to 1, 1)\}$. (And $Y_{3}=Y_{2}\cup \{1\}.$)
\end{itemize}
Notice that the case $c_1$ obtained is exactly case $\tilde{c}_1$ in Example \ref{ex:CasebaseCanavotto}. If we consider $c_{2}= (X_2, op(s_{2}), f(s_{2})=0)$, we obtain the case $\tilde{c}_2$ in Example \ref{ex:CasebaseCanavotto}.  
\end{example}

Notice that Definition~\ref{ex:opinionState} mimic Definition~\ref{def:OpinionCase}. Indeed, we can verify that a case obtained from a state is indeed a case in the sense of Definition~\ref{def:cb}, as shown in the following proposition. 
\begin{proposition} \label{prop:ocb}
Let $s=(X, sol)\in S$ such that $f(s)\neq {\sf ?}$. Given $c=(X, op(s), f(s))$ obtained from $s$, it is a case in Definition~\ref{def:cb}: (1) $op(s)$ is an opinion, and (2) $f(s)$ is the outcome supported by $op$.
\end{proposition}

 We can now show how a classifier model can be translated into a case base. This can be done by collecting all the cases obtained from the assessed states in the classifier model.
\begin{definition}
Let $C=(S,f)$ a RCM. The case base obtained from $C$ is defined as $CB_{C}= \{c=(X, op(s), f(s)) \mid s \in S  \textrm{ and } f(s)\neq ?\}$.
\end{definition}

\begin{example}\label{ex:casebasestate}
Consider $C_{ex}$ from Example~\ref{ex:classifier}. Then $CB_{C_{ex}}=\{c_1, c_2\}$, where $c_{1}=(X_1, op(s_{1}), f(s_{1}))$  and $c_{2}=(X_2, op(s_{2}), 0)$ (see Example~\ref{ex:opinionState}). But then $CB_{ex}$ is exactly $\tilde{\cb}$ in Example~\ref{ex:CasebaseCanavotto}. So $\cb$  is consistent because $\tilde{\cb}$ is consistent (see Example~\ref{ex:CasebaseCanavotto2}).
\end{example}

\subsection{Constraining New Cases}

In this subsection, we want to show how the output for new cases provided by a classifier should be constrained by relevant cases. In particular, we want to define the relevance relation between states that can capture the precedential constraint in \cite{Canavotto1}. To do so, we will need some further terminology.

\subsubsection{Relevance Relation Modelling Precedential Constraint}

\begin{definition}[Factors in a state favoring a side of a concern]\label{def:factsState}
Let $H$ be a hierarchy, $m=degree(0/1)$, and $s=(X, sol)\in S$. We define $\tilde{s}= X\cup \bigcup_{degree(t/\overline{t})< m} \{Conclusion(rule_{sol}(t/\overline{t}))\}$. Given $t\in Int\cup Dec$, we also define $Facts^{t}(s)\defin \tilde{s} \cap Facts^{t}$.
\end{definition}

The set $\tilde{s}$ contains the base situation of $s$ and all the conclusions of the rules in the solution $sol$ of $s$, except the rules pertaining the issue $0/1$. Hence, $\tilde{s}$ contains both base factors of $s$ and intermediate factors that apply to $s$, namely those intermediate factors supported by the solution $sol$ in $s$. In this sense, $Facts^{t}(s)$ is the set of factors in $s$ favoring side $t$ wrt the concern $t/\overline{t}$, while $Facts^{\overline{t}}(s)$ is the set of factors in $s$ favoring side  $t/\overline{t}$ wrt the concern $t/\overline{t}$.

\begin{example}
Consider $C_{ex}$ of Example \ref{ex:classifier}, with a focus on $s_{1}$. So $\tilde{s_{1}}= \{f_2, f_3, f_4, f_5, p, r, q\}$. Given the concern $q/q'$, the factor in $s_1$ favoring $q$ is the intermediate factor $p$, namely $Facts^{q}(s_{1})= \tilde{s_{1}}\cap Facts^{q}= \{p\}$. While, the factor in $s_1$ favoring the opposite side $q'$ is the base factor $f_3$, namely $Facts^{q'}(s_1)=\{f_3\}$. 
In contrast, since the solution $sol_{3}$ of $s_{3}$ is $\emptyset$, $\tilde{s_{3}}=X_{3}$: in $\tilde{s_3}$ there are only the base factor already in $s_3$.  
\end{example}

\begin{definition}[Reason and decision for a concern]\label{def:ReasForConcern}
Let $s=(X, sol)\in S$, $u/\overline{u} \in Concern$, and $t \in \{u, \overline{u}\}$. Then:  
\begin{itemize}
     \item $dec^{u/\overline{u}} (s) = Conclusion(r)$ if exists $r \in Rules$ such that $r = rule_{sol}(u/\overline{u})$, otherwise $dec^{u/\overline{u}} (s) = ?$; 
    \item $Reas^{t}(s) = Antecedent(r)$ if exists $r =rule_{sol}(u/\overline{u})$ such that $t=Conclusion(r)$, otherwise $Reas^t(s) = \emptyset$. 
\end{itemize}
\end{definition}

Intuitively, $Reas^{t}(s)$ is the reason for $dec^{u/\overline{u}}(s) =t$, which is the decision in $s$ wrt the concern $u/\overline{u}$, if such decision exists.

\begin{example}
Consider $C_{ex}$ from Example~\ref{ex:classifier} with a focus on the concern $p/p'$ and $s_{1}$. Now $\tilde{s_{1}}= \{f_2, f_3, f_4, f_5, p, r, q\}$. We know $\{f_2\}\to p\in sol_{1}$. It gives us $Reas^{p}(s_{1})=\{f_2\}$ and $Reas^{p'}(s_{1})=\emptyset$, together with $dec^{p/p'}(s_{1})=p$. Consider $s_{3}$. Now $Reas^{p}(s_{3})=\emptyset$, due to the solution of $s_{3}$ is empty and $dec^{p/p'}(s_{3})=?$ --- which means that $s_{3}$ has not been decided on the concern $p/p'$.    
\end{example}

We can now define a relevance relation that allows us to capture the precedential constraint by~\cite{Canavotto1}.

\begin{remark}
Let $u/\overline{u} \in Concern$, $t\in \{u,\overline{u}\}$, and $K\subseteq \atm_{0}\cup Int\cup Dec$. So $K^t \defin K \cap Facts^t$ includes all the factors in favor of $t$ in $K$. 
\end{remark}

\begin{definition}[Concern-based Relevance]\label{def:concernRelevance}
Given a classification function $f$ and $u/\overline{u} \in Concerns$, we define $\mathcal{R}^{u/\overline{u}} \subseteq 2^{\atm_{0}\cup  Int} \times 2^{\atm_{0}\cup  Int} \times S \times S$ as a relevance relation, when it satisfies: $(D,G, s, s') \in \mathcal{R}^{u/\overline{u}}$ iff $Reas^{t}(s)\subseteq Reas^{t}(s') \cup D^{t}$ and $Reas^{\overline{t}}(s') \cup G^{\overline{t}}\subseteq Facts^{\overline{t}}(s)$, where $dec^{u/\overline{u}}(s)=t$ and $f(s) \neq ?$.
\end{definition}

\begin{example} \label{ex:relevance1} The definition of relevance provided above offers a computational method to verify precedential constraints between two states~\cite{Horty2011RR, Canavotto1}. In other words, it helps determine whether one state should be (or should have been) assessed based on the decision made for the other state, following a form of \emph{a fortiori} reasoning. This verification can be done by fixing specific $D$ and $G$, which in turn depends on the decisions associated with the states. We will start by considering two states decided in opposite directions. Consider $s_{1}$ as in the Example~\ref{ex:classifier}. Let $s_{4}= (X_{4}, sol_{4})$ with $f(s_{4})=0$,  
$X_{4}= \{f_1, f_2, f_3, f_4, f_5\}$ and $sol_{4}= \{\{f_3\}\to p’, \{f_5\}\to  r’  , \{p’\}\to q’, \{r’\}\to 0\}$. We focus on the concern $p/p'$. Note that, with respect to $p/p'$, the states $s_1$ and $s_4$ were decided in opposite ways: $dec^{p/p'}(s_1)= p$, while $dec^{p/p’}(s_4)= p'$. 
Notice now that: 
\begin{itemize}
    \item We have $(Facts^{p}(s_{4}), \emptyset, s_{1}, s_{4}) \in \mathcal{R}^{p/p'}$, because
    \begin{itemize}
        \item[(a)] $Reas^{p}(s_1) \subseteq  Facts^{p}(s_{4})$ and 
        \item[(b)] $Reas^{p'}(s_4) \subseteq Facts^{p'}(s_1)$;
    \end{itemize}
        
        \item However, $( \emptyset, Facts^{p}(s_{4}), s_{1}, s_{4}) \not\in \mathcal{R}^{p/p'}$, because
        \begin{itemize}
        \item[(c)] although $Facts^{p'}(s_4) \subseteq Facts^{p'}(s_1)$, 
        \item[(d)] we have $Reas^{p}(s_1) \not\subseteq  Reas^{p}(s_{4})=\emptyset$. 
    \end{itemize}
\end{itemize}

    Statement $(a)$ tells us that $Reas^{p}(s_1)$, the  reason for $p$ in $s_{1}$, is contained in the factors favoring $p$, in $s_{4}$. Hence, it could have been used in $s_{4}$. On the other hand, according to $(b)$,  the reason for $p'$ in $s_{4}$, $Reas^{p'}(s_4)$,  is contained in the factors favoring $p'$, in $s_{1}$; so it could have been used in $s_{1}$. But $s_{1}$, was decided for $p$. Thus, in deciding $s_{1}$ the reason $Reas^{p}(s_1)$ for $p$ was considered stronger than the reason $Reas^{p'}(s_4)$ for $p'$.  Therefore, --- supposing that $s_4$ was decided later than $s_{1}$~\footnote{We will return to the importance of the time dimension in Section~\ref{section:binding} later.} --- $s_4$, should have been decided as $s_1$, \emph{a fortiori}. 
    
    What has just been discussed is nothing new: if we see it from the perspective of case base models \cite{Canavotto1}, we have simply said that $s_{4}$ has violated the precedential relation established by $s_{1}$ --- that is, the decision for $s_{4}$ is inconsistent with the decision for $s_{5}$.~\footnote{In other terms, $s_{1}$ is a ``conflicting case'' wrt the decision in $s_{4}$, as extensively discussed in~\cite{CanavottoPrecedential}.} Here, our relevance relation provides a specific instance to compute this inconsistency. Indeed, we have the following result.

\end{example}

\begin{proposition}\label{prop:equivalnceInconsistency}
Let $u/\overline{u} \in Concern$. Given $s=(X, sol), s' =(X', sol')\in S$, where $f(s)\neq ?$ and $f(s')\neq ?$, we can define $CB=\{c, c'\}$ with $c =(X, op(sol), f(s))$ and $c'= (X', op(sol'), f(s'))$. Then the following conditions are equivalent: 
\begin{enumerate}
    \item $(Facts^{t}(s'), \emptyset, s, s') \in \mathcal{R}^{u/\overline{u}}$, where $dec^{u/\overline{u}}(s)=t$ and \\$dec^{u/\overline{u}}(s')=\overline{t}$;
    \item $CB$ is inconsistent wrt $u/\overline{u}$.
\end{enumerate}
\end{proposition}
\begin{example}
    Consider again Example~\ref{ex:relevance1}. Let us now consider $c_{1}$ and $c_{4}$ the cases obtained from $s_{1}$ and $s_{4}$, as well as $\cb= \{c_{1}$, $c_{4}\}$. We know, from the previous example, that $dec^{p/p'}(s_{1})=p$, $dec^{p/p'}(s_{4})=p'$ and $(Facts^{p}(s_{4}), \emptyset, s_{1}, s_{4}) \in \mathcal{R}^{p/p'}$. Then, by Proposition~\ref{prop:equivalnceInconsistency}, $\cb$ is inconsistent. We can verify this result as follows. By Definition~\ref{casefromstate}, the decisions related to $p/p'$ in $c_{1}$ and $c_{4}$ are respectively $d_{1}= (X_{1}, r_{1}, p)$ and $d_{4}= (X_{4}, r_{4}, p')$. From $c_{1}$, we retrieve $\{f_{3}\}<_{c_1} \{f_{1}, f_{2}\}$.  On the other hand, from $c_{4}$, we retrieve $\{f_{1}, f_{2}\}<_{c_4} \{f_{3}\}$. Hence, $\cb$ is inconsistent.  
\end{example}

\subsubsection{Classifications by Relevance: From Precedents to New Cases}

From the previous section we know that we can express the precedential constraint between decided states (in the opposite direction) on the basis of the relevance relation. We would like to do the same for undecided states, i.e. we would like to express via the relevance relation, when a new state $s^{*} $, is to be decided in a certain direction with respect to a concern, based on the precedential constraint. To do this, however, we must consider some aspects. Firstly, in a new case we only know the base factors, and not the intermediate factors.  And solely on the basis of the base factors, we cannot always apply the precedential constraint to decide on concerns of degree greater than $1$ (such as the top issue $0/1$). What we can do is to apply a precedential constraint ‘conditioned’ by the presence of certain intermediate factors in a set $F$, i.e. we can say that if certain intermediate factors in a set $F$ applied in $s^{*}$, then $s^{*}$ would be constrained to be decided in a certain way on the basis of some previous cases. Can $F$ be any set of facts? No, in the sense that we require $F$ to be factors that can be obtained from the factual situation in $s^{*}$, i.e. for which there exists a solution $sol$ applicable to the factual situation in $s^{*}$ and such that the rules in $sol$ support the factors in $F$. In this sense, $F$ is obtainable from $s^{*}$. 

\begin{definition}\label{def:compatible}
Let $F\subseteq Int$ and $X\subseteq \atm_{0}$. We say that $F$ is obtainable from $X$ if there is a solution $sol\in Sol$, such that
\begin{itemize}
\item $sol$ is applicable to $X$;
\item $\forall p\in F$: $\exists r \in sol$ such that $Conclusion(r)=p$.
\end{itemize}
In this case, we say that $sol$ is compatible with $F$ and $X$. We denote $Sol_{F,X}$ the set of all $sol\in Sol$ compatible with $F$ and $X$. 
\end{definition}

However, it may happen that a solution compatible with $F$ may have rules that support more intermediate factors than those in $F$. Given a specific concern $u/\overline{u}$, we will  require that the solution compatible with $F$ is minimal with respect to $u/\overline{u}$ , i.e. no more intermediate factors favoring $u/\overline{u}$ than those in $F$ and in the factual situation of $s^{*}$ are introduced by the solution. 

\begin{definition}\label{def:iminimal}
Let $F\subseteq Int$, $X\subseteq \atm_{0}$, and $u/\overline{u}\in Concern$. Given $sol\in Sol_{F,X}$ a compatible solution with $F$ and $X$, we say that $sol$ is $u/\overline{u}$-minimal wrt $F$ and $X$ iff, for any $r=rule_{sol}(u/\overline{u})$, it is the case that $Antecedent(r) \subseteq (Facts^{u}\cup Facts^{\overline{u}})\cap (X\cup F)$.
\end{definition}

Namely, $sol$ is $u/\overline{u}$-minimal wrt $F$ and $X$ if $sol$ doesn't introduce more factors favoring $u$ and $\overline{u}$ wrt the ones in  $F\cup X$.

 Here is where the relevance relation and the sets $D$ and $G$ in the relevance relation comes into play. 
  
In general, when we write $(D ,G, s, s^{*}) \in \mathcal{R}^{u/\overline{u}}$, we mean that $s$ is relevant to $s^{*}$, given the presence of $D$ and $G$.  
Also, notice that by Definition~\ref{def:concernRelevance}, the following two statements are equivalent: given $dec^{u/\overline{u}}(s) = t$ and $f(s^{*})=?$ (which leads to $Reas^{t}(s^{*})=\emptyset$), 
\begin{enumerate}
    \item $(D, G, s, s^{*}) \in \mathcal{R}^{u/\overline{u}}$;
    \item both $Reas^{t}(s)\subseteq D^{t}$ and $G^{t} \subseteq Facts^{\overline{t}}(s)$. 
\end{enumerate}

Which sets $D$ and $G$ are we interested in now? We consider $F$ a set obtainable from $s^{*}$.   
So $D$ and $G$ under consideration are now  respectively, $F^{t}\cup  Facts^{t}(s^{*})$ (the factors in favor of $t$ in $F$ and   $s^{*}$)  and $F^{t}\cup  Facts^{\overline{t}}(s^{*})$ (the factors in favor of $\overline{t}$ in $F$ and $s^{*}$). Suppose $(F^{t}\cup  Facts^{t}(s^{*}),F^{\overline{t}}\cup  Facts^{\overline{t}}(s^{*}), s, s^{*}) \in \mathcal{R}^{u/\overline{u}}$, with $dec^{u/\overline{u}}(s) = t$.
So we know that $Reas^{t}(s)\subseteq F^{t} \cup  Facts^{t}(s^{*})$  and $F^{\overline{t}} \cup  Facts^{t}(s^{*}) \subseteq Facts^{\overline{t}}(s)$.
These two conditions are crucial. Why? Take any compatible and minimal solution $sol'$ wrt $F$ and the factual situation $X^{*}$ of $s^{*}$, we can define a new state $s'=(X^{*}, sol')$.

Now, $Reas^{\overline{t}}(s')\subseteq F^{\overline{t}} \cup  Facts^{t}(s^{*}) $, whatever the solution wrt $t/\overline{t}$ for $s'$ is. Indeed, we have two possibilities. Solution $sol'$ for $s'$ satisfies $dec^{u/\overline{u}}(s')= t$ (that is, $s'$ is decided as $t$ given $sol'$). So $Reas^{\overline{t}}(s')=\emptyset$ and trivially $Reas^{\overline{t}}(s')\subseteq F^{\overline{t}} \cup  Facts^{t}(s^{*}) $. If, instead, solution $sol'$ for $s'$ satisfies $dec^{u/\overline{u}}(s')= \overline{t}$, the minimality of $sol'$ will ensure $Reas^{\overline{t}}(s')\subseteq F^{\overline{t}} \cup  Facts^{t}(s^{*})$ --- because $sol'$ cannot introduce more factors favoring $\overline{t}$ than the ones in $F$ and $s^{*}$. 

On the other hand, the compatibility of solution ensures that $F^{t} \cup  Facts^{t}(s^{*}) \subseteq  Facts^{t}(s')$ --- $s'$ and $s^{*}$ have the same factual situation $X^{*}$ and, by the compatibility,  every element in $F$ is the conclusion of some rule in $sol'$.

Combining  the conditions obtained so far, we can have that $Reas^{t}(s)\subseteq Facts^{t}(s')$ and $Reas^{\overline{t}}(s')\subseteq Facts^{\overline{t}}(s)$, which leads to the main result: $(Facts^{t}(s' ), \emptyset, s, s' ) \in \mathcal{R}^{u/\overline{u}}$. 

Now we summarize. From $(F^{t}\cup  Facts^{t}(s^{*}),F^{\overline{t}}\cup  Facts^{\overline{t}}(s^{*}), s, s^{*}) \in \mathcal{R}^{u/\overline{u}}$, we can retrieve $(Facts^{t}(s' ), \emptyset, s, s' ) \in \mathcal{R}^{u/\overline{u}}$ for any new state $s'$ obtained by enriching $s^{*}$ with a minimal compatible solution. So, by Proposition~\ref{prop:equivalnceInconsistency} we have that, if $s'$ is decided differently than $s$ wrt $u/\overline{u}$ then $s'$ is inconsistent wrt $s$;  and viceversa, if $s'$ is inconsistent wrt $s$ then $s'$ is decided differently than $s$ wrt $u/\overline{u}$. Differently stated, $(F^{t}\cup  Facts^{t}(s^{*}),F^{t'}\cup  Facts^{t'}(s^{*}), s, s^{*}) \in \mathcal{R}^{u/\overline{u}}$ tell us that if we choose a  solution for $s^{*}$ that  does not add further elements wrt $F$ and $X^{*}$ pertaining $u/\overline{u}$ (i.e. it is minimal) we will introduce incosistency wrt $s$, pertaining $u/\overline{u}$, only if the solution  goes against $s$ wrt the decision for $u/\overline{u}$. 
This intuitions are behind the following proposition.

\begin{proposition}\label{prop:equivalnceInconsistency2}
   Given $ u/\overline{u}\in Concern$, we define $s=(X,sol)\in S$ such that $f(s)\neq ?$ and  $dec^{u/\overline{u}}(s)= t \in \{u,\overline{u}\}$ and define $s^{*}=(X^{*},sol^{*})\in S$ such that $f(s^{*})=?$ and $sol^{*}=\emptyset$. Given $F\subseteq Int$, we also define $F^{t}= F\cap Facts^{t}$ and $F^{\overline{t}}= F\cap Facts^{\overline{t}}$. Now, given $sol'\in Sol_{F,X^{*}} $ such that $sol'$ is $u/\overline{u}$-minimal wrt $F$ and $X^{*}$, we define $s'=(X^{*}, sol')$. Let $CB=\{c, c'\}$ be with $c =(X, op(sol), f(s))$ and $c'= (X', op(sol'), f(s'))$.   
    
    Assume $(F^{t}\cup Facts^{t}(s^{*}), F^{\overline{t}}\cup Facts^{\overline{t}}(s^{*}) , s, s^{*}) \in \mathcal{R}^{u/\overline{u}}$. Then:
    \begin{align*}
        dec^{u/\overline{u}}(s')=\overline{t} \Leftrightarrow CB \text{ is incosistent wrt }u/\overline{u}.
    \end{align*}
\end{proposition}

Given $s^{*}\in S$ a new case we want to define a way to infer a decision for the new case. In this sense, we want to infer a decision making process for assessing the case.

\begin{definition}[Decision making process/function wrt a concern] Given $u/\overline{u}\in Concern$ and $s^{*}\in S$ with $f(s^{*})\neq ?$, a \emph{decision making process/function for $s^{*}$ wrt $u/\overline{u}$ is any function $k^{*}: S \to 2^{\{u,\overline{u}\}}$ that satisfies: 
\begin{equation*}
k^{*}(s) =
        \begin{cases}
\{f(s)\}  &\text{ if } s\neq s^{*};\\ 
V \in 2^{\{u,\overline{u}\}} &\text{ otherwise. }  
    \end{cases}
\end{equation*}}
\end{definition}

We define now a decision making process for the top issue $0/1$, that is iteratively constructed on decision making processes for intermediate concerns $u/\overline{u}$. These processes are defined on the basis of the relevance relation we extracted from the precedential contraint in \cite{Canavotto1}. 

\begin{definition}[Decision-making Process]\label{def:Relevantdecisionmaking}
Assume that $s^{*}= (X^{*}, \emptyset)\in S$ with $f(s^{*})=?$. Given $m=degree(0/1)$, we iteratively define a decision making process for $s^{*}$, based on the  relevant states, $h^{*}: S \to 2^{\{0,1\}}$, as follows. 
\begin{itemize}
\item Define $F_{0}= X^{*}$. 
\item Let $n=1,...,m$. Define $u/\overline{u}\in Concern$ with $degree(u/\overline{u})= n$.
\begin{itemize}

\item For $t\in \{u, \overline{u}\}$, define $cite^{t}({s^{*}})\defin \{s \mid ( F^{t}_{n-1}, F^{\overline{t}}_{n-1},s, s^{*})  \in \mathcal{R}^{u/\overline{u}}  \}$; that is, $cite^{t}({s^{*}})$ is the set of relevant states for $s^{*}$ that can be cited to constrain a decision for $s^{*}$ as $t$. 
\item Define the decision making process related to $u/\overline{u}$ as a function $h_{u/\overline{u}}^{*}: S \to 2^{\{u, \overline{u}\}}$ such that: for all $s\in S$, 
\begin{equation*}
h_{u/\overline{u}}^{*}(s) \defin
        \begin{cases}
\{dec^{u/\overline{u}}(s')\} &\text{ if } s\neq s^{*};\\ 
\{t \mid cite^{t}({s^{*}})\neq \emptyset\} &\text{ otherwise. }    
    \end{cases}
\end{equation*}

\item Define $F_{n}\defin F_{n-1}\cup \bigcup_{u/\overline{u}:degree(u/\overline{u})=n} h^{*}_{u/\overline{u}}(s^{*}) $. 

For $t\in Int\cup Dec$, define $F^{t}_{n}\defin F^{t}\cap F_{n} $.
\end{itemize}
\end{itemize}

\end{definition}

Notice that at each step $n$, the set $F_{n}$ collects all base factor in $X^{*}$ (since they are in $F_{0}$) along with all conclusions from the decision making process, pertaining to concerns up to degree $n$. We define the set of base and intermediate factors and  final decisions determined for $s^{*}$ by the decision making process as $F_{s^{*}}\defin F_{m}$.

\begin{example}
We consider the classifier  in Example \ref{ex:classifier}. We know $degree(0/1)=3$.
Consider $s_{3}$. Recall that $X_{3}= \{f_1, f_2, f_3, f_4, f_5, f_6\}$. 
\begin{itemize}
\item let $n=0$ 
\begin{itemize}
\item  $degree(p/p')=1$. It can be verified that the set of   factors that support $p$ at degree $0$ is $F_{0}^{p}=\{f_1,f_2\}$ and the set of  factors against $p$ is $F_{0}^{p'}=\{f_3\}$. So, it implies $(F_{0}^{p}, F_{0}^{p'}, s_{1}, s_{3}) \in \mathcal{R}^{p/p'} $ --- considering the support/against factors, we can say that $s_1$ is relevant for $s_3$ wrt concern $p/p'$. Also, it holds $(F_{0}^{p}, F_{0}^{p'},  s_{2}, s_{3}) \in \mathcal{R}^{p/p'}$ --- considering the support/against factors, we can say that $s_2$ is  relevant for $s_3$ wrt concern $p/p'$. So $cite^{p}(s_{3})=\{s_1, s_{2}\}$, namely  $s_1$ and $s_2$ can be a cited to constrain the decision as $p$ for $s_3$. Hence, our decision making process outputs  $h_{p/p'}^*(s_{3}) = \{p\}$.
\item  $degree(r/r')=1$, $F_{0}^{r}=\{f_5,f_6\}$, and $F_{0}^{r'}=\{f_4\}$.  It can be verified that $(F_{0}^{r'},F_{0}^{r},  s_{2}, s_{3}) \in \mathcal{R}^{r/r'} $ as well as  $(F_{0}^{r}, F_{0}^{r'},s_{1},s_{3}) \not \in\mathcal{R}^{r/r'} $. So $cite^{r'}(s_{3})=\{s_2\}$. Hence, $h_{r/r'}^*(s_{3}) = \{r'\}$.
\item So, we can define $F_{1}$ as $F_{0}\cup \bigcup_{u/\overline{u}:degree(u/\overline{u})=1} h^{*}_{u/\overline{u}}(s_{3}) =X_{3}\ \cup \{p, r'\}$ according to the decision making process. 
\end{itemize}
\item let $n=2$.
\begin{itemize}
\item  $degree(q/q')=2$. So we have $F_{1}^{q}=\{p\}$ and $F_{1}^{q'} =\emptyset$. It can be verified that $(F_{1}^{q},F_{1}^{q'},  s_{1}, s_{3}) \in \mathcal{R}^{q/q'} $ and $(F_{1}^{q},F_{1}^{q'}, s_{2}, s_{3}) \in \mathcal{R}^{q/q'} $. It turns out  $cite^{q}(s_{3})=\{s_1, s_2\}$. So, $h_{q/q'}^*(s_{3}) =\{q\}$.
\item We then define $F_{2}$ as $ F_{1}\cup \bigcup_{u/\overline{u}:degree(u/\overline{u})=2} h^{*}_{u/\overline{u}}(s_{3}) =F_{1}\ \cup \{q\}= X_{3}\cup \{p,r',q\}$.  
\end{itemize}
\item let $n=3$.
\begin{itemize}
\item $degree(0/1)=3$. So it follows that $F_{2}^{1}=\{q\}$ and $F_{2}^{0} =\{r'\}$. Then we get $(F_{2}^{1}, F_{2}^{0}, s_{2}, s_{3}) \in \mathcal{R}^{0/1} $, however we also have $(F_{2}^{0},F_{2}^{1}, s_{1}, s_{3}) \not \in \mathcal{R}^{0/1} $. So, it follows that $cite^{0}(s_{3})=\{s_2\}$. Hence $f_{0/1}^*(s_{3}) = \{0\} $. 
\item We then define $F_{3}$ as $F_{2}\cup \bigcup_{u/\overline{u}:degree(u/\overline{u})=3} h^{*}_{u/\overline{u}}(s_{3}) =F_{2}\ \cup \{0\}= X_{3}\cup \{p,r',q, 0\}$. 
\end{itemize}
\item The set of factors and final decision for $s_{3}$ is then $F_{s_3}=F_{3} =X_{3}\cup \{p,r',q, 0\}$.
\end{itemize}
\end{example}

On the basis of the cardinality of the decision outcomes related to a concern, we can determine when the decision making process in unambiguous with respect to the concern.  

\begin{definition} \label{def:ambiguious}
Let $u/\overline{u}\in Concern$. Let $k^{*}: S \to 2^{\{u, \overline{u}\}}$ be a decision making process related to $u/\overline{u}$. We say that: 
\begin{itemize}
    \item No decision can be made for $s^{*}$ wrt $u/\overline{u}$ iff  $k^{*}(s^{*})=\emptyset$;
    \item $k^{*}$ is unambiguous for $s^{*}$ iff $|k^{*}(s^{*})| = 1$; 
    \item $k^{*}$ is ambiguous for $s^{*}$ iff $|k^{*}(s^{*})|>1$.
\end{itemize}
\end{definition}
\noindent Thus, a decision making function $k^{*}$  related to $u/\overline{u}$ is ambiguous for $s^{*}$ iff $k^{*}(s^{*})= \{u, \overline{u}\} $, i.e. when  $s^{*}$ can be decided both as $u$ and as $u'$ (namely, we have a conflict situation).

Consider the starting case base obtained from $s_1, s_2$ in the previous example. This case base is consistent --- see Example~\ref{ex:casebasestate}. We can actually prove that if the starting case base is consistent, then the decision making process is unambiguous both for top-level issue and for all intermediate concerns. This follows from a specific property of the relevance relation: if two cases decided for  opposite sides wrt a concern are relevant given two sets $F$ and $G$, for the same case, then, they are relevant one for the other. Namely, we have the following result.

\begin{proposition}\label{prop:euclideanRelevance}
Let $u/\overline{u}\in Concern$. Given $s_1, s_2, s_0\in S$, we assume $dec^{u/\overline{u}}(s_1)= u$ and $dec^{u/\overline{u}}(s_2)= \overline{u}$. Let $F, G\subseteq \atm_{0}\cup Int$. So: 
\begin{align*}
    &\textrm{If } (F, G ,s_1, s_0)  \in \mathcal{R}^{u/\overline{u}} \textrm{ and } ( G, F ,s_2, s_0)  \in \mathcal{R}^{u/\overline{u}}, \\
    &\textrm{then }(Facts^{u}(s_2), \emptyset ,s_1, s_2)  \in \mathcal{R}^{u/\overline{u}} \textrm{ and }(Facts^{\overline{u}}(s_1), \emptyset ,s_2, s_1)  \in \mathcal{R}^{u/\overline{u}}. 
\end{align*}
\end{proposition}

On the basis of the previous proposition we can verify that, if the  case base obtained from a classifier model is consistent, then the decision making process is unambiguous.

\begin{proposition}\label{prop:ConsAmb}
Let $C=(S,f)$ be a RCM with $s^{*}\in S$ such that $f(s^{*})=?$. Let $\cb$ be the case base obtained from $C=(S,f)$. If $\cb$ is consistent, then $h_{u/\overline{u}}^{*}$ is not ambiguous for  $s^{*}$, where $u/\overline{u}\in Concern$.
\end{proposition}

The viceversa does not hold: we can obtain a decision making process that is unambiguous with respect to to the top-level issue, even though the case base is inconsistent with respect to some concern (and so, inconsistent). Consider the following example.

\begin{example}\label{ex:IncNotAmbiguous}
Let $S \subseteq2^{Atm_0}\times Sol$. Consider $X=\{f_1,f_2,f_3, f_4, f_5\}$ and the following  states:
\begin{itemize}
    \item $s_{6}=(X, sol_6)$ with $sol_{6}=\{\{f_1\}\to p, \{p\}\to q, \{f_5\}\to r', \{q\}\to 1 \}$;
    \item $s_{7}= (X, sol_7)$ with $sol_{7}=\{ \{f_1\}\to p, \{p\}\to q,  \{f_4\}\to r, \{q\}\to 1  \}$.
    \item $s_{*}= (X, sol_{*})$ with  $sol_{*}= \emptyset$. 
\end{itemize}
Let $f: 2^{Atm_0}\times Sol \to \val $ be defined as: $f(s)=1$ iff $s\in \{s_6, s_7\}$; $f(s)=?$, otherwise. Consider the cases $c_{6}=(X, op(s_{6}), 1)$ and $c_{7}=(X,op(s_{7}), 1 )$, which are obtained from $s_{6}$ and $s_{7}$, respectively.  The case base $CB=\{c_6, c_7\}$ is inconsistent --- Because $\{f_4\}<_{c_6}^{r/r'} \{f_5\}$ and $\{f_5\}<_{c_7}^{r/r'} \{f_4\}$. Now consider the decision making process:
\begin{itemize}
\item let $n=0$: 
\begin{itemize}
\item $degree(p/p')=1$. So $F_{0}^{p}=\{f_1,f_2\}$ and $F_{0}^{p'}=\{f_3\}$, which imply $(F_{0}^{p}, F_{0}^{p'}, s_{6}, s^{*}) \in \mathcal{R}^{p/p'} $ and  $(F_{0}^{p}, F_{0}^{p'}, s_{7}, s^{*}) \in \mathcal{R}^{p/p'} $. It follows $cite^{p}(s^{*}) = \{s_6, s_7\}$. So, $h_{p/p'}^*(s^{*}) = \{p\}$.
\item $degree(r/r')=1$. 
Similarly, we can compute to have $cite^{r'}(s^{*})=\{s_6\}$ and $cite^{r}(s^{*})=\{s_7\}$ --- we retrieve two constraining precedents in conflict wrt $r/r'$. In such a case, it results $h_{r/r'}^*(s^{*}) = \{r,r'\}$. 
\item We thus define $F_{1}$ as $F_{0}\cup \bigcup_{u/\overline{u}:degree(u/\overline{u})=1} h^{*}_{u/\overline{u}}(s^{*}) =X\ \cup \{p, r, r'\}$, due to the decision making process. 
\end{itemize}
\item let $n=2$:
\begin{itemize}
\item  $degree(q/q')=2$. 
We can compute to have $cite^{q}(s^{*})=\{s_6, s_7\}$. Hence $h_{q/q'}^*(s^{*}) = \{q\}$. 
\item We then define $F_{2}$ as $F_{1}\cup \bigcup_{u/\overline{u}:degree(u/\overline{u})=2} h^{*}_{u/\overline{u}}(s^{*}) =F_{1} \cup \{q\}  =X \cup \{p,r,r',q\}$.
\end{itemize}
\item let $n=3$:
\begin{itemize}
\item $degree(0/1)=3$. 
We can compute to have $cite^{1}(s^{*})=\{s_6\}$. Hence it is the case that $h_{0/1}^*(s^{*}) =\{1\}$. 
\end{itemize}
\item  The set $F_{s^{*}}$ of intermediate factors for $s^{*}$ is equal to $F_{3}=  X\cup \{p,r',r,q,1\}$. 
\end{itemize}

\end{example}

Let us take a closer look at the previous example. If we consider the concern $r/r'$, we have that, given $s_6$, the new state $s^{*}$ should be decided as $r'$, while given $s_7$ it should be decided as $r$. For all the other issues we have no problems: $s^{*}$ should be decided as $p$ and as $q$ on the basis of both $s_6$ and $s_7$. When we come to decide for the top-level issue, we realise that the decision regarding the concern $r/r'$ is negligible 
for the overall decision: In both situations, $s_6$ is relevant for $s^{*}$ and, on the basis of it, $s^{*}$ should be decided for $1$. In other words, let us see the situation from the perspective of priority ordering induced by the cases. Suppose we choose $r'$ for $s^{*}$. From $s_6$, we know that $q$ is a stronger reason for $1$ than the reason $r'$ for $0$, so $s^{*}$ should be decided for $1$, to maintain consistency with $s_6$. Suppose, instead, we choose $r$. Then we must decide $s^{*}$ for $1$, either in virtue of $s_6$ or in virtue of $s_7$ (or much more trivially because in $s^{*}$ there are only factors in favor of 1). Following what was just discussed, we introduce the following definition. 

\begin{definition}
Let $s^{*}\in S$ be $f(s^{*})=?$. Given $u/\overline{u}\in Concern$, we say that $u/\overline{u}$ is negligible  with respect to decision for $s^{*}$, pertaining the top issue $0/1$, iff   $|h^{*}_{0,1}(s^{*})|=1$  and $h^{*}_{u,\overline{u}}$ is ambiguous for $s^{*}$.
\end{definition}

In other words, $u/\overline{u}$ is negligible  with respect to the decision for $s^{*}$, if, although the decision about $u/\overline{u}$ is ambiguous, the decision for 0/1 can be made unambiguously. 

We have a decision making process to work out which intermediate factor and which top issue should be chosen for the new state $s^{*}$, on the basis of relevant states. In the situation that the decision making process leads to a non-ambiguous solution for the top-level issue (i.e. $|h^{*}_{0,1}(s^{*})|=1$),  which solution can be chosen for $s^{*}$? Intuitively, any solution that is compatible with   the set of base factors and intermediate factors in $F_{s^{*}}$ and is $u/\overline{u}$-minimal, for all concerns raised by $F_{s^{*}}$, could be chosen. The problem we have, however, is that, as we saw in the previous example, we may have that for a concern $u/\overline{u}$ both $u$ and $\overline{u}$ are in $F_{s^{*}}$ --- in this sense we have a conflict in $F_{s^{*}}$. In this situation, no solution that is compatible with $F_{s^{*}}$ can be found (remember that in a solution there can be at most one rule for each concern). We must therefore partition $F_{s^{*}}$ into subsets containing, for each concern $u/\overline{u}$, either $u$ or $\overline{u}$. 

\begin{definition}[Conflict-free partition]\label{def:compatiblePartition}
Let $s^{*}\in S$, where $f(s^{*})=?$ and $|h^{*}_{0,1}(s^{*})|=1$, and $F_{s^{*}}$ be the final set  determined by the decision making process. 
We define the conflict-free partition $\mathcal{P}$ of $F_{s^{*}}$ as the set $\{P\subseteq F^{*}_{s^{*}}\mid  |P\cap \{u, \overline{u}\}|=1 \text{ for all } u/\overline{u}\in  Concern \text{ such that } \{u, \overline{u}\}\cap F_{s^{*}}\neq \emptyset \}$.
\end{definition}

Notice that when $F_{s^*}$ is not empty, we have $\mathcal{P}$ to be non empty.

\begin{example}\label{ex:partitions}
Consider the Example~\ref{ex:IncNotAmbiguous}. We have that $\mathcal{P}=\{P_{1}, P_{2}\} $, where $P_{1}= X\cup \{p,q,r,1\}$ and $P_{2}= X\cup \{p,q,r',1\}$. 
\end{example}

On the basis of the concept of partition, we can define a decision process that provides us with admissible solutions for the case under consideration, given the starting classifier model. We want this decision making process to be in line, as much as possible, with the precedential constraint. Specifically, we seek to avoid introducing new inconsistencies beyond those already present in the classifier model (i.e., in the derived case base). To do this, we must ensure that the rules contained in the solution for $s^{*}$ are compatible with the rules already present in the solutions of the relevant states for $s^{*}$.  The condition \ref{cond:sol} in the following definition is for this purpose: in the new solution, we enter a rule pertaining a concern, only if the antecedent of that rule is stronger than (i.e. contains) the reason for the decision pertaining to that concern of a case relevant to $s^{*}$.  

\begin{definition}[Decision making process for a solution]\label{def:decisionSol}
Let $s^{*}=(X^{*}, \emptyset)\in S$ be with $|h^{*}_{0,1}(s^{*})|=1$. The \emph{decision making process/function for a solution for $s^{*}$, based on relevants states}, is the function $g^*: S \to 2^{Sol}$ defined as follows:
\begin{equation*}
g^{*}(s) =
        \begin{cases}
\{sol\}   &\text{ if } s=(X,sol)\neq s^{*};\\ 
\{sol_{P}\mid P\in \mathcal{P} \} & \text{ otherwise.} 
    \end{cases}
\end{equation*}
where $sol_{P}\in Sol_{P\cap Int, X^{*}}$ must satisfy this condition: for all $t\in P$ 
\begin{equation} \tag{$\star$}\label{cond:sol}
\begin{split}
r: Antecedent(r)\to t \in sol_{P} \text{ iff } &\\Reas^{t}(s)\subseteq Antecedent(r)\subseteq P^{t},
&\text{ for some }s\in cite^{t}(s^{*})
\end{split}
\end{equation}
\end{definition}

\begin{example}
Consider the previous Examples~\ref{ex:IncNotAmbiguous} and \ref{ex:partitions}. Recall  $X=\{f_1,f_2,f_3, f_4, f_5\}$,
    $\mathcal{P}=\{P_{1}, P_{2}\} $, where $P_{1}= X\cup \{p,q,r,1\}$. We show how to obtain solutions for $s^{*}$ compatible with $P_{1}$. 
\begin{itemize}
\item Since $p\in P_{1}$ and $s_{6}, s_{7}\in cite^{p}(s^{*})$, we have to take  rules  $r_{p}$ related to the concern $p/p'$ s.t.\\
$r_{p}: Antecedent(r_{p})\to p$ and  \\
$Reas^{p}(s_6)= Reas^{p}(s_{7})= \{f_{1}\}\subseteq Antecedent(r_{p})\subseteq \{f_{1}, f_{2}\}$.
\item Since $r\in P_{1}$ and  $ s_{7}\in cite^{r}(s^{*})$,  we have to take  rules  $r_{r}$ related to the concern $r/r'$ s.t.\\
$r_{r}: Antecedent(r_{p})\to p$ and  \\
$Reas^{r}(s_7)=  \{f_{4}\}\subseteq Antecedent(r_{r})\subseteq \{f_{4}\}$. Hence the unique rule we can take is $r_{r}:\{f_{4}\}\to r$.
\item  We know $q\in P_{1}$. The unique rule  $r_{q}$ related to the concern $q/q'$ we can take is $r_{q}:\{p\}\to q$.
\item Since $1\in P_{1}$ and $ s_{6}\in cite^{1}(s^{*})$,  we have to take  rules  $r_{1}$ related to the concern $0/1$ s.t.\\
$r_{1}: Antecedent(r_{1})\to 1$ and  \\
$Reas^{1}(s_6)=  \{f_2\}\subseteq Antecedent(r_{q})\subseteq \{q,r\}$. 
\end{itemize}
The same reasoning pattern can be used to find rules compatible wrt $P_2$.
\end{example}
 We can prove that if the starting case base is consistent, then the decision making process so defined will provide us with at least a solution --- Hence, $g^{*}(s^{*}) \neq \emptyset$. 
  
\begin{proposition}\label{prop:partition}
Let $C=(S,f)$ be a classifier model such that $s^{*}=(X^{*}, \emptyset)\in S$ with $|h^{*}_{0,1}(s^{*})|=1$. Let $\cb$ be the case base obtained from $C$. If $\cb$ is consistent, then: 
\begin{itemize}
\item [I)] $g^{*}(s^{*})\neq \emptyset$;
\item [II)] for all $sol_{P}\in g^{*}(s^{*})$,  $Conclusion(rule_{sol_{p}}(0/1)) = h_{0/1}^*(s^*)$; 
\item [III)] $ sol \in g^*(s^*)$ is $u/\overline{u}$ minimal.
\end{itemize}
\end{proposition}

Consider the previous Example \ref{ex:partitionsrules}, in addition with $s'=(X^{*}, sol')$. Let $c'=(X^{*}, op(sol'),  f_{0/1}^{*}(s^{*}))$, where $sol'\in g^{*}(s^{*})$ and $f_{0/1}^{*}(s^{*})=1$.  Consider the case base $CB=\{c_6, c_7\}$ in Example~\ref{ex:IncNotAmbiguous}. This was inconsistent. Then, $CB\cup \{c'\}$ will also be inconsistent. Instead, if the starting case base is consistent, adding the new decided case, we will still have a consistent case base.

\begin{proposition}\label{prop:solConsistency}
Let $s^{*}\in S$ be with $f(s^{*})=?$ and $|h^{*}_{0,1}(s^{*})|=1$, and $\cb$ the case obtained from $C=(S,f)$. If $\cb$ is consistent and $c'=(X^{*}, op(sol'),  f_{0/1}^{*}(s^{*}))$, then $\cb\cup \{c'\}$ is still consistent.
\end{proposition}

\section{Importance of Courts and Time for Conflict Resolution}\label{section:binding}
The decision making process on the basis of relevance is not always unambiguous. This also happens because the model seen so far is a flat model \emph{with respect to cases}: all cases are equally important. This is not always the case in the legal domain: some cases are more important than others, because they have been decided, for instance, by a higher court or later in time. Furthermore, not all cases are to be considered binding: binding cases for a case at hand are the relevant cases that were decided earlier and by a court that has binding power on the court deciding the case at hand. Lastly, some binding cases are subject to exceptions: either because they have been overruled (i.e. there is a relevant case decided later in a different manner, by a court with the power to do so) or because they have been decided \emph{per incuriam} (i.e. ignoring a binding precedent themselves). These insights are formalised in~\cite{PrecedentsClash}, where the classifier models in~\cite{LiuLorini2021BCL} are enriched with the court structure of the legal system under consideration and a temporal dimension among cases. In~\cite{PrecedentsClash} a single-step relevance relation is take into account. Without going into technical details, we now provide an example of how the framework in \cite{PrecedentsClash} can be used in what has been done so far, to resolve ambiguities in the decision making process.

\begin{example}
Suppose we have  a simple legal system in which we have three courts $k_0$, $k_1$, $k_2$. Court $k_0$ is a higher court than $k_1$, which in turn is a higher court than $k_2$. Let us assume that \emph{vertical stare decisis} applies (higher courts always express decisions that are binding on lower courts). 
Moreover, a higher court can always overrule a lower court (so $k_{0}$ can overrule both $k_1$ and $k_2$), while a lower court can never overrule a higher court (for instance, $k_{2}$ cannot overrule either $k_1$ or $k_0$). 
Finally, suppose that $k_2$ is self bound. 
We consider a classifier model such that

\begin{itemize}
\item $f(s_{8})=1$,  $s_{8}=(X_{8}, sol_{8})$, where $X_{8}= \{f_1,f_3\}$ and $sol_{8}= \{\{f_{1}\}\to p, \{p\}\to q, \{q\}\to 1\}$;
\item $f(s_{9})=0$,  $s_{9}=(X_{9}, sol_{9})$, where $X_{9}= \{f_1,f_2,f_3, f_4, f_5\}$ and $sol_{9}= \{\{f_{3}\}\to p', \{f_{5}\}\to r' ,  \{p'\}\to q', \{r'\}\to 0\}$;
\item $f(s_{10})=1$,  $s_{10}=(X_{10}, sol_{10})$, where $X_{10}= \{f_3, f_4,f_5\}$ and $sol_{10}= \{\{f_{3}\}\to p', \{p'\}\to q', \{f_4\}\to r,\{r\}\to 1 \}$.
\end{itemize}

Suppose that $s_{8}$ has been assessed by court $k_1$, $s_{9}$ has been assessed by court $k_0$, and  $s_{10}$ has been assessed by court $k_2$. In addition, $s_{8}$ has been assessed before  $s_{9}$ and $s_{9}$ has been assessed before $s_{10}$. 

Consider a new case $s^{*}=(X_{*}, \emptyset)$, where $X_{*}=\{f_1, f_2,f_3, f_4, f_5\}$. $s^{*}$ is to be decided by the lowest court $k_{2}$. 
Without any consideration of time or courts, we could not make an unambiguous decision for $s^{*}$. Indeed, $s_{9}$ and $s_{8}$ are both relevant for $s^{*}$ with respect to $p/p'$, but in different directions. Also, $s_{9}$ and $s_{10}$ are relevant for $s^{*}$ with respect to $r/r’$, but in different directions. And the decisions with respect to $p/p'$ and $r/r’$ are both non-negligible for the overall decision. In this sense, we have a conflict of precedents. 
However, if we consider the time element and the hierarchy of courts, we can resolve the conflict of precedents for $s^{*}$. Notice that
\begin{itemize}
\item $s_8$ was relevant for $s_9$ wrt concern $p/p'$. However, $s_9$ was decided differently than $s_8$, by a court, $k_0$, with overruling power wrt the court of $s_8$, namely $k_1$. Hence, $s_9$ overruled $s_8$ wrt $p/p'$. We write $Overruled_{p/p'}(s_{8})$. Notice also that $s_8$ was not relevant for $s_9$ wrt $r/r'$. In this sense, we have a \emph{partial overruling}: $s_9$ overruled $s_8$, only with respect to a specific concern. 
\item $s_9$ was  decided before $s_{10}$ and was relevant and binding for $s_{10}$ --- $k_{0}$ is higher than $k_{2}$ --- wrt the concern $r/r'$. However, $s_{10}$ was decided differently than $s_{9}$, by the court $k_2$ without overruling power wrt the court of $s_{9}$, namely $k_0$. Hence, $s_{10}$ was decided \emph{per incuriam} wrt $r/r'$. We write $Incuriam_{r/r'}(s_{10})$. As before, only one part of the decision in $s_{10}$ is \emph{per incuriam}, the one pertaining $r/r'$. 
\end{itemize}
Now, we consider the binding precedents for $s^{*}$ without exceptions (i.e. those neither overruled nor to be \emph{per incuriam}). Recall that the court in $s^{*}$ is  $k_2$, which is bound by  $k_{0}$, $k_{1}$ and $k_{2}$. Notice that:
\begin{itemize}
\item The only binding precedent  without exceptions for $s^{*}$, pertaining $p/p'$ is $s_{9}$; because $s_{8}$ has been overruled. So $s^{*}$ should follow $s_{9}$ and be decided as $p'$. And, on the basis of that, $s^{*}$ should be decided as $q'$.
\item The only binding precedent  without exceptions for $s^{*}$, pertaining $r/r'$ is $s_{9}$; because $s_{10}$ is \emph{per incuriam}. So $s^{*}$ should follow $s_{9}$ and be decided as $r'$.
\item So we know that $p',q'$, and $r'$ should apply in $s^{*}$. Hence, $s^{*}$ should be decided as $0$, on the basis of the binding precedent $s_{9}$ (or simply because all factors in $s^{*}$ are for $0$). 
\end{itemize}

\end{example}
We conclude this section with an observation. Proposition~\ref{prop:euclideanRelevance} states that, if two cases are in conflict regarding the case under consideration --- meaning they are both relevant for the case under consideration but have been decided in the opposite direction --- then they must be relevant to each other. The previous example might suggest that, in such conflicts, one of the two conflicting cases is either overruled or decided \emph{per incuram} --- and therefore one of the two conflicting cases is  not actually
binding. However, this is not always true. 
The above example has simplified certain concepts, particularly the  notion of \emph{per incuram}. For instance, as discussed in~\cite{PrecedentsClash}, a case decided by a lower court can never ignore a precedent decided by a higher court, even if \emph{per incuram}.~\footnote{For example, in the  case Cassell v Broome (UK), it was affirmed that the Court of Appeal could not disregard a House of Lords decision, even though  \emph{per incuriam} \cite{interpreting}. }
Hence, it may happen that a case  \emph{per incuram} remains binding for the case at hand. The impact of these nuanced considerations on our classifier framework will be explored in future research.

\section{Conclusions} \label{sec:conclusion}
In this paper, we provide a first example of how classifier models in the legal domain can be enriched to  take also into account sets of rules used to decide cases (potentially both the \emph{ratio decidendi} of the case and the rules established by the precedents constituents). We have provided an initial example of how this can be done, starting from the framework proposed in \cite{Canavotto1}.

Some issues deserve further attention.
First of all, we would like to highlight additional similarities   between our work and the work in \cite{Canavotto1,CanavottoPrecedential}. Specifically, our concern-based relevance relation  between cases decided in opposite ways  returns exactly the definition of conflicting cases introduced in \cite{CanavottoPrecedential}, in the context of the flat reason model. This is because, like the notion of conflicting cases in \cite{CanavottoPrecedential}, our relevance relation for cases is extracted from the definition of inconsistent case base.
\cite{CanavottoPrecedential} defines when a potential decision for a new case 
is obligatory, even in the context of an inconsistent case base. This happens  when there is a conflicting case for the opposite decision and no supporting case for the  opposite decision exists.
These two last condition are equivalent to requiring that the decision making process related to a single concern --- introduced in Definition \ref{def:Relevantdecisionmaking} --- is not ambiguous. 
For this reason, we suppose that the decision making process based on the relevance relation in our definition leads to the same result that would be obtained if the notion of obligatory decision in \cite{CanavottoPrecedential} were applied in the multi-step framework in \cite{Canavotto1}.
The main difference between our work in \cite{CanavottoPrecedential}, lies in the fact that we provide a way to compute which rules are admissible for a new case, something that,  to the best of our understanding, is not explicitly defined in  \cite{Canavotto22}.

As already mentioned, the work presented here is a first step within a larger project. We aim to import the hierarchical and temporal elements, as defined in \cite{PrecedentsClash}, into the framework of classifier models described in this paper.
This will allow us to refine the decision making process in Definition \ref{def:Relevantdecisionmaking}. As outlined in Section~\ref{section:binding}, only the binding precedents without exceptions --- determined through courts and time ---  should be considered when constraining a new case. 
In addition, in the enriched framework, we can also introduce hierarchical and temporal principles for resolving conflicts between binding precedents.

\section*{Acknowledgements}
Cecilia Di Florio is partially supported by ISA doctoral Prize, issued by UNIBO ISA DP. Huimin Dong is partially funded by the WWTF project nr. 10.47379/ICT23-030. 
Antonino Rotolo was partially supported by the projects CN1 “National Centre for HPC,
Big Data and Quantum Computing” (CUP: J33C22001170001) and PE01 “Future Artificial Intelligence Research” FAIR (CUP: J33C22002830006).

\bibliographystyle{ACM-Reference-Format}
\bibliography{sample-base}


\begin{thebibliography}{18}


\ifx \showCODEN    \undefined \def \showCODEN     #1{\unskip}     \fi
\ifx \showDOI      \undefined \def \showDOI       #1{#1}\fi
\ifx \showISBNx    \undefined \def \showISBNx     #1{\unskip}     \fi
\ifx \showISBNxiii \undefined \def \showISBNxiii  #1{\unskip}     \fi
\ifx \showISSN     \undefined \def \showISSN      #1{\unskip}     \fi
\ifx \showLCCN     \undefined \def \showLCCN      #1{\unskip}     \fi
\ifx \shownote     \undefined \def \shownote      #1{#1}          \fi
\ifx \showarticletitle \undefined \def \showarticletitle #1{#1}   \fi
\ifx \showURL      \undefined \def \showURL       {\relax}        \fi
\providecommand\bibfield[2]{#2}
\providecommand\bibinfo[2]{#2}
\providecommand\natexlab[1]{#1}
\providecommand\showeprint[2][]{arXiv:#2}

\bibitem[Aleven(1997)]%
        {CATO}
\bibfield{author}{\bibinfo{person}{Vincent A. W. M.~M. Aleven}.} \bibinfo{year}{1997}\natexlab{}.
\newblock \emph{\bibinfo{title}{Teaching case-based argumentation through a model and examples}}.
\newblock \bibinfo{thesistype}{Ph.\,D. Dissertation}. \bibinfo{address}{USA}.
\newblock
\showISBNx{0591729873}
\newblock
\shownote{AAI9821228}.


\bibitem[Bench-Capon(2023)]%
        {BenchCapon2}
\bibfield{author}{\bibinfo{person}{Trevor Bench-Capon}.} \bibinfo{year}{2023}\natexlab{}.
\newblock \showarticletitle{A Note on Hierarchical Constraints}. In \bibinfo{booktitle}{\emph{JURIX 2023}}.
\newblock
\showISBNx{9781643684727}
\urldef\tempurl%
\url{https://doi.org/10.3233/FAIA230940}
\showDOI{\tempurl}


\bibitem[Bench-Capon and Atkinson(2021)]%
        {BenchCapon1}
\bibfield{author}{\bibinfo{person}{Trevor Bench-Capon} {and} \bibinfo{person}{Katie Atkinson}.} \bibinfo{year}{2021}\natexlab{}.
\newblock \showarticletitle{Precedential constraint: the role of issues}. In \bibinfo{booktitle}{\emph{Proceedings of ICAIL 21}}. \bibinfo{publisher}{Association for Computing Machinery}, \bibinfo{pages}{12–21}.
\newblock
\showISBNx{9781450385268}
\urldef\tempurl%
\url{https://doi.org/10.1145/3462757.3466062}
\showDOI{\tempurl}


\bibitem[Branting(1991)]%
        {branting}
\bibfield{author}{\bibinfo{person}{L.~Karl Branting}.} \bibinfo{year}{1991}\natexlab{}.
\newblock \showarticletitle{Reasoning with portions of precedents}. In \bibinfo{booktitle}{\emph{Proceedings of ICAIL '91}}. \bibinfo{publisher}{ACM}.
\newblock
\urldef\tempurl%
\url{https://doi.org/10.1145/112646.112664}
\showDOI{\tempurl}


\bibitem[Bruninghaus and Ashley(2003)]%
        {IBP}
\bibfield{author}{\bibinfo{person}{Stefanie Bruninghaus} {and} \bibinfo{person}{Kevin~D. Ashley}.} \bibinfo{year}{2003}\natexlab{}.
\newblock \showarticletitle{Predicting outcomes of case based legal arguments}. In \bibinfo{booktitle}{\emph{Proceedings of ICAIL 03}} \emph{(\bibinfo{series}{ICAIL '03})}. \bibinfo{publisher}{ACM}, \bibinfo{pages}{233–242}.
\newblock
\showISBNx{1581137478}
\urldef\tempurl%
\url{https://doi.org/10.1145/1047788.1047838}
\showDOI{\tempurl}


\bibitem[Canavotto(2022)]%
        {Canavotto22}
\bibfield{author}{\bibinfo{person}{Ilaria Canavotto}.} \bibinfo{year}{2022}\natexlab{}.
\newblock \showarticletitle{Precedential Constraint Derived from Inconsistent Case Bases}. In \bibinfo{booktitle}{\emph{{JURIX} 2022}}. \bibinfo{publisher}{{IOS} Press}.
\newblock
\urldef\tempurl%
\url{https://doi.org/10.3233/FAIA220445}
\showDOI{\tempurl}


\bibitem[Canavotto(2023)]%
        {CanavottoPrecedential}
\bibfield{author}{\bibinfo{person}{Ilaria Canavotto}.} \bibinfo{year}{2023}\natexlab{}.
\newblock \showarticletitle{Reasoning with inconsistent precedents}.
\newblock \bibinfo{journal}{\emph{Artificial Intelligence and Law}} (\bibinfo{year}{2023}).
\newblock
\showISBNx{1572-8382}
\urldef\tempurl%
\url{https://doi.org/10.1007/s10506-023-09382-7}
\showDOI{\tempurl}


\bibitem[Canavotto and Horty(2023a)]%
        {Canavotto2}
\bibfield{author}{\bibinfo{person}{Ilaria Canavotto} {and} \bibinfo{person}{John Horty}.} \bibinfo{year}{2023}\natexlab{a}.
\newblock \bibinfo{booktitle}{\emph{The Importance of Intermediate Factors}}.
\newblock
\showISBNx{9781643684727}
\urldef\tempurl%
\url{https://doi.org/10.3233/FAIA230941}
\showDOI{\tempurl}


\bibitem[Canavotto and Horty(2023b)]%
        {Canavotto1}
\bibfield{author}{\bibinfo{person}{Ilaria Canavotto} {and} \bibinfo{person}{John Horty}.} \bibinfo{year}{2023}\natexlab{b}.
\newblock \showarticletitle{Reasoning with hierarchies of open-textured predicates}.
\newblock
\showISBNx{9798400701979}


\bibitem[Di~Florio et~al\mbox{.}(2023)]%
        {lail}
\bibfield{author}{\bibinfo{person}{Cecilia Di~Florio}, \bibinfo{person}{Xinghan Liu}, \bibinfo{person}{Emiliano Lorini}, \bibinfo{person}{Antonino Rotolo}, {and} \bibinfo{person}{Giovanni Sartor}.} \bibinfo{year}{2023}\natexlab{}.
\newblock \showarticletitle{Finding Factors in Legal Case-Based Reasoning}. In \bibinfo{booktitle}{\emph{Logics for AI and Law (LAIL-23)}}. \bibinfo{publisher}{College Publications}.
\newblock


\bibitem[Florio et~al\mbox{.}(2024)]%
        {PrecedentsClash}
\bibfield{author}{\bibinfo{person}{Cecilia~Di Florio}, \bibinfo{person}{Huimin Dong}, {and} \bibinfo{person}{Antonino Rotolo}.} \bibinfo{year}{2024}\natexlab{}.
\newblock \showarticletitle{When Precedents Clash}.
\newblock In \bibinfo{booktitle}{\emph{JURIX 2024}}. \bibinfo{publisher}{IOS Press}, \bibinfo{pages}{34 -- 47}.
\newblock
\urldef\tempurl%
\url{https://doi.org/doi:10.3233/FAIA241232}
\showDOI{\tempurl}


\bibitem[Florio et~al\mbox{.}(2023)]%
        {FlorioLLRS23}
\bibfield{author}{\bibinfo{person}{Cecilia~Di Florio}, \bibinfo{person}{Xinghan Liu}, \bibinfo{person}{Emiliano Lorini}, \bibinfo{person}{Antonino Rotolo}, {and} \bibinfo{person}{Giovanni Sartor}.} \bibinfo{year}{2023}\natexlab{}.
\newblock \showarticletitle{Inferring New Classifications in Legal Case-Based Reasoning}. In \bibinfo{booktitle}{\emph{Proceedings of {JURIX} 2023}}. \bibinfo{publisher}{{IOS} Press}.
\newblock
\urldef\tempurl%
\url{https://doi.org/10.3233/FAIA230942}
\showDOI{\tempurl}


\bibitem[Horty(2011)]%
        {Horty2011RR}
\bibfield{author}{\bibinfo{person}{John~F. Horty}.} \bibinfo{year}{2011}\natexlab{}.
\newblock \showarticletitle{Rules and reasons in the theory of precedent}.
\newblock \bibinfo{journal}{\emph{Legal theory}}  \bibinfo{volume}{17} (\bibinfo{year}{2011}), \bibinfo{pages}{1--33}.
\newblock
\urldef\tempurl%
\url{https://doi.org/10.1017/S1352325211000036}
\showDOI{\tempurl}


\bibitem[LexisNexis(2004)]%
        {LexisNexis2}
\bibfield{author}{\bibinfo{person}{LexisNexis}.} \bibinfo{year}{2004}\natexlab{}.
\newblock \bibinfo{title}{Glossary, https://www.lexisnexis.co.uk/legal/glossary/ratio-decidendi}.
\newblock
\newblock
\urldef\tempurl%
\url{https://www.lexisnexis.co.uk/legal/glossary/ratio-decidendi}
\showURL{%
\tempurl}


\bibitem[Liu and Lorini(2021)]%
        {LiuLorini2021BCL}
\bibfield{author}{\bibinfo{person}{Xinghan Liu} {and} \bibinfo{person}{Emiliano Lorini}.} \bibinfo{year}{2021}\natexlab{}.
\newblock \showarticletitle{A logic for binary classifiers and their explanation}. In \bibinfo{booktitle}{\emph{CLAR 2021}}. Springer.
\newblock


\bibitem[Liu and Lorini(2023)]%
        {LiuLoriniJLC}
\bibfield{author}{\bibinfo{person}{Xinghan Liu} {and} \bibinfo{person}{Emiliano Lorini}.} \bibinfo{year}{2023}\natexlab{}.
\newblock \showarticletitle{A unified logical framework for explanations in classifier systems}.
\newblock \bibinfo{journal}{\emph{Journal of Logic and Computation}} \bibinfo{volume}{33}, \bibinfo{number}{2} (\bibinfo{year}{2023}), \bibinfo{pages}{485--515}.
\newblock


\bibitem[Liu et~al\mbox{.}(2022)]%
        {liu2022modelling}
\bibfield{author}{\bibinfo{person}{Xinghan Liu}, \bibinfo{person}{Emiliano Lorini}, \bibinfo{person}{Antonino Rotolo}, {and} \bibinfo{person}{Giovanni Sartor}.} \bibinfo{year}{2022}\natexlab{}.
\newblock \showarticletitle{Modelling and Explaining Legal Case-Based Reasoners Through Classifiers}.
\newblock In \bibinfo{booktitle}{\emph{JURIX 2022}}. \bibinfo{publisher}{IOS Press}, \bibinfo{pages}{83 -- 92}.
\newblock
\urldef\tempurl%
\url{https://doi.org/10.3233/FAIA220451}
\showDOI{\tempurl}


\bibitem[MacCormick and (editors)(1997)]%
        {interpreting}
\bibfield{author}{\bibinfo{person}{MacCormick} {and} \bibinfo{person}{Summers (editors)}.} \bibinfo{year}{1997}\natexlab{}.
\newblock \bibinfo{booktitle}{\emph{Interpreting Precedents A Comparative Study}}.
\newblock \bibinfo{publisher}{Ashgate}.
\newblock


\end{thebibliography}

\end{document}